\documentclass[conference]{IEEEtran}
\IEEEoverridecommandlockouts
\usepackage{cite}
\usepackage{amsmath,amssymb,amsfonts}
\usepackage{amsthm}
\usepackage{subfigure}
\usepackage{algorithm}
\usepackage{algorithmic}
\usepackage{graphicx}
\usepackage{textcomp}
\usepackage{url}
\usepackage{xcolor}
\DeclareMathOperator*{\argmax}{arg\,max}

\def\BibTeX{{\rm B\kern-.05em{\sc i\kern-.025em b}\kern-.08em
    T\kern-.1667em\lower.7ex\hbox{E}\kern-.125emX}}
\begin{document}

\title{Active Learning for Video Classification with Frame Level Queries\\

\thanks{This research was supported in part by the National Science Foundation under Grant Number: 2143424}
}


\author{\IEEEauthorblockN{Debanjan Goswami}
\IEEEauthorblockA{\textit{Department of Computer Science} \\
\textit{Florida State University}\\
}
\and
\IEEEauthorblockN{Shayok Chakraborty}
\IEEEauthorblockA{\textit{Department of Computer Science} \\
\textit{Florida State University}\\
}
}



\maketitle

\begin{abstract}
Deep learning algorithms have pushed the boundaries of computer vision research and have depicted commendable performance in a variety of applications. However, training a robust deep neural network necessitates a large amount of labeled training data, acquiring which involves significant time and human effort. This problem is even more serious for an application like video classification, where a human annotator has to watch an entire video end-to-end to furnish a label. Active learning algorithms automatically identify the most informative samples from large amounts of unlabeled data; this tremendously reduces the human annotation effort in inducing a machine learning model, as only the few samples that are identified by the algorithm, need to be labeled manually. In this paper, we propose a novel active learning framework for video classification, with the goal of further reducing the labeling onus on the human annotators. Our framework identifies a batch of exemplar videos, together with a set of informative frames for each video; the human annotator needs to merely review the frames and provide a label for each video. This involves much less manual work than watching the complete video to come up with a label. We formulate a criterion based on uncertainty and diversity to identify the informative videos and exploit representative sampling techniques to extract a set of exemplar frames from each video. To the best of our knowledge, this is the first research effort to develop an active learning framework for video classification, where the annotators need to inspect only a few frames to produce a label, rather than watching the end-to-end video. Our extensive empirical analyses corroborate the potential of our method to substantially reduce human annotation effort in applications like video classification, where annotating a single data instance can be extremely tedious. 
\end{abstract}


\section{Introduction}
\label{sec_intro}

With the widespread deployment of modern sensors and cameras, images and videos have become ubiquitous. This has encouraged the development of video classification algorithms to analyze their semantic content for various applications, such as search, summarization, security and surveillance among others. Deep neural networks (CNN and LSTM architectures) have depicted commendable performance in this field \cite{Review_Paper}. Common methods include obtaining global video-level descriptors using CNN architectures \cite{Ng_2015}, processing videos at two spatial resolutions: a low-resolution context stream and a high-resolution fovea stream \cite{Karpathy_Paper}, fusion technique to integrate data representations at the frame level and video level \cite{Tian_2019} among others. However, for all these models to work reliably, a large amount of labeled training data is essential, gathering which is an expensive process in terms of time, labor and human expertise. Thus, an algorithm to reduce the human labeling effort is of immense importance in video classification applications. 

\begin{figure*}[ht]
	\centering
		\subfigure[Conventional AL query for video classification]{
          \label{fig_val_conv}
          \includegraphics[width=.4\textwidth]{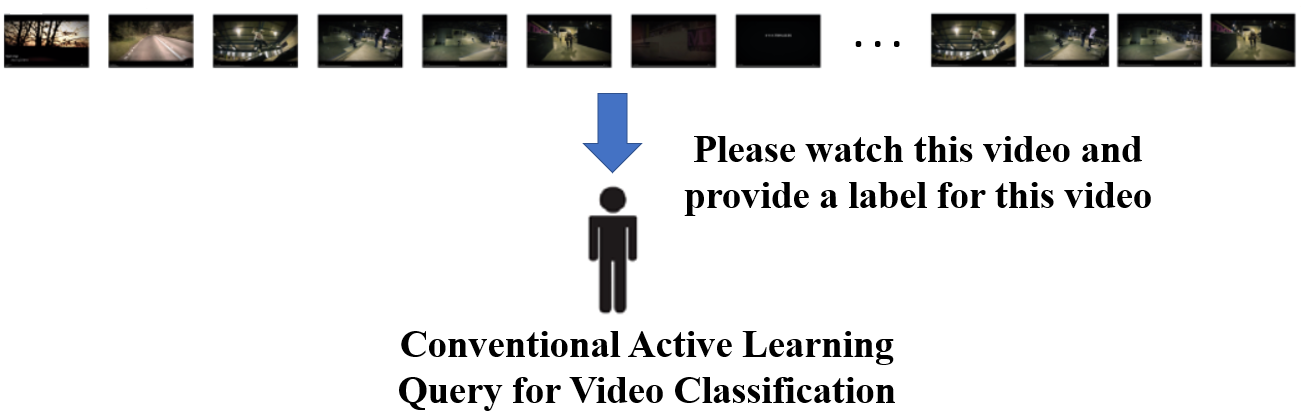}}
     \hspace{.03in}
     \subfigure[Proposed active query and annotation mechanism]{
          \label{fig_val_proposed}
          \includegraphics[width=.4\textwidth]{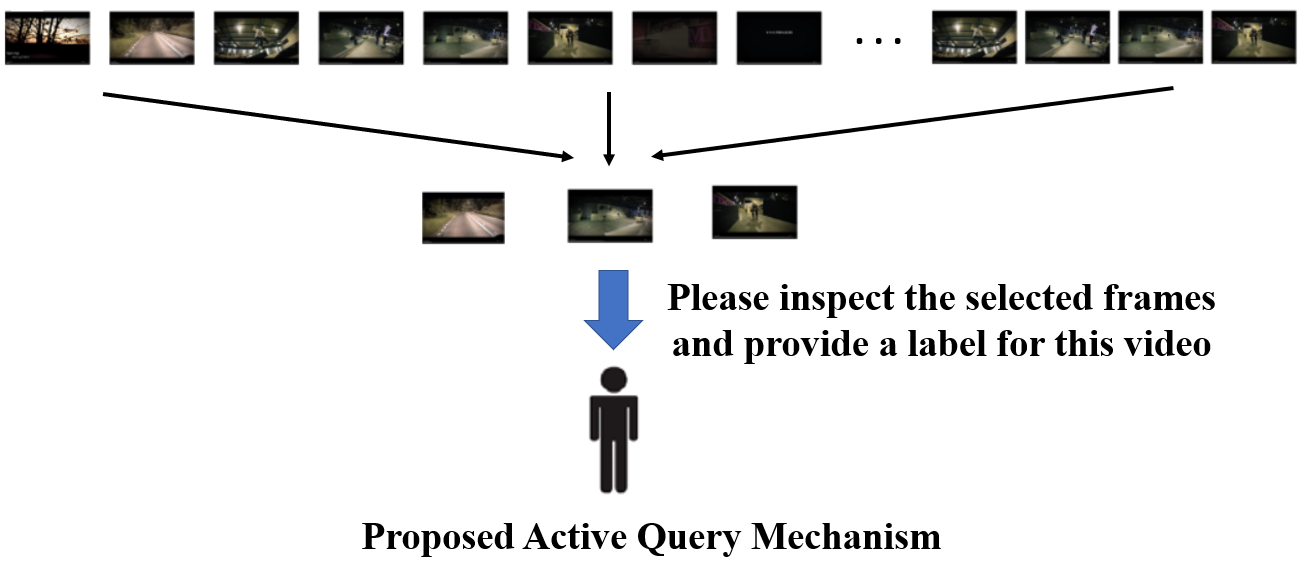}}
    
      \caption{(a) Conventional AL query for video classification, where the human annotator has to watch a video end-to-end to provide a label. (b) Proposed active query and annotation mechanism where we actively select a batch of videos, together with a subset of frames from each video; the human annotator merely needs to inspect the frames and provide a label for the video.}
	\label{fig_motivation}
\end{figure*} 

\textit{Active Learning (AL)} is a machine learning paradigm, where the goal is to automatically identify the salient and exemplar samples from large amounts of redundant data \cite{Settles_2010}. This  tremendously reduces the human annotation effort in inducing a machine learning model, since the human expert only has to label the samples queried by the algorithm. Further, since the model gets trained on the exemplar samples from the data population, it typically depicts better generalization performance than a model where the training data is selected at random. This is an extremely relevant paradigm in today's world, where an enormous amount of digital data is being generated, but there is a serious dearth of human labor to annotate the data to induce learning models. AL has been successfully used in a variety of applications, including computer vision \cite{Yoo_2019}, text analysis \cite{tong_support_2000}, computational biology \cite{Hatice_2010} and medical diagnosis \cite{Gorriz_2017} among others. Active learning is particularly relevant in the context of deep learning, in order to reduce human annotation effort in training the data-hungry deep neural networks \cite{Ren_2021}. 

Designing an AL algorithm for a video classification application entails the human annotator to meticulously watch each queried video end-to-end in order to furnish a label \footnote{we use the terms \textit{annotators}, \textit{oracles}, \textit{labelers} and \textit{users} synonymously in this paper}. This is an extremely time-consuming and laborious process; the annotators may get bored and fatigued quickly and lose interest in the task. This necessitates specialized and more user-friendly query and annotation mechanisms, to utilize the available human labor more efficiently. In this paper, we propose a novel active learning algorithm to address this challenging and practical problem. Our algorithm identifies a batch of informative videos, together with a set of exemplar frames from each; the human annotator merely has to review the queried frames and furnish a label for each video. This is illustrated in Figure \ref{fig_motivation}. Providing such feedback is significantly less time-consuming and burdensome than watching an end-to-end video. We formulate an optimization problem based on an uncertainty and diversity based criterion to identify a batch of informative videos, and exploit representative sampling techniques to select a subset of exemplar frames from each. To our knowledge, this is the first active learning framework for video classification which poses label queries based on a set of exemplar frames, rather than the complete video. We hope this research will motivate the development of AL algorithms with other novel annotation mechanisms, with the goal of further reducing the labeling burden on human oracles in a video classification application. 

The rest of the paper is organized as follows: we present a survey of related research in Section \ref{sec_related}, our active sampling framework is detailed in Section \ref{sec_proposed}, the results of our empirical studies are presented in Section \ref{sec_expt}, and we conclude with discussions in Section \ref{sec_conc}.


\section{Related Work}
\label{sec_related}

In this section, we present an overview of active learning in general, followed by a survey of AL for video classification. 

\textbf{Active Learning}: AL has received significant research attention in the machine learning community. Uncertainty sampling is by far the most common strategy for active learning, where unlabeled samples with highest classification uncertainties are queried for their labels. The uncertainty of an unlabeled sample can be computed by its entropy \cite{Holub_2008}, its distance from the separating hyperplane in the feature space for SVM classifiers \cite{tong_support_2000}, the disagreement among a committee of classifiers regarding the label of the sample \cite{freund_selective_1997}, the expected error reduction of the future learner \cite{Fu_KDD_2018} and so on. Submodular optimization techniques have also been exploited for active data sampling \cite{AL_Submod_1, AL_Submod_2}. 

The growing success and popularity of deep learning has motivated research in the field of deep active learning (DAL), where the goal is to select informative unlabeled samples to efficiently train a deep neural network \cite{Ren_2021}. A task agnostic AL framework was proposed by Yoo and Kweon \cite{Yoo_2019} that incorporated a loss prediction module in the network architecture, to predict the loss value of an unlabeled sample and query samples accordingly. A DAL framework based on core-set selection was proposed by Sener and Savarese \cite{coreset_AL}, which selected a batch of samples, such that the deep model trained on the selected samples depicts similar performance to that trained on the whole dataset. DAL has also been studied in conjunction with neural architecture search \cite{Geifman_2019}, which queries samples for labeling and simultaneously searches for the best neural architectures on-the-fly. A novel training loss function for DAL was proposed by Shui \textit{et al.}, where active sample selection and traning the network parameters were achieved through alternating optimization \cite{Shui_2020}. Deep active learning techniques based on adversarial training have depicted particularly impressive performance \cite{Ducoffe_2018, Mayer_2020, Zhang_2020, Sinha_2019}. Active learning has also been studied in conjunction with other learning paradigms such as transfer learning \cite{Chattopadhyay_2013}, reinforcement learning \cite{AL_RL} etc. Moreover, the idea of identifying an informative set of samples for human inspection has been extended to other problem domains, such as matrix completion \cite{Ruchansky_2015}, video summarization \cite{Molino_2017} and feature selection \cite{Shim_2018} among others. 

Recently, there have been efforts to design AL systems with novel query and annotation mechanisms, with the goal of further reducing the labeling burden on human annotators. Joshi \textit{et al.} \cite{Joshi_2010} proposed a binary query mechanism which queried an unlabeled sample together with a potential class label and the user had to provide the binary answer as to whether the queried unlabeled sample belonged to the selected class or not. Along similar lines, Biswas and Jacobs proposed an AL algorithm for clustering, which queried a pair of samples and the oracles needed to specify whether or not the samples in a pair correspond to the same cluster \cite{Biswas_2012}. Xiong \textit{et al.} \cite{Xiong_2015} proposed a triplet query framework to learn approximate distance metrics for a nearest neighbor classifier; the algorithm queried unlabeled data triplets $(x_{i}, x_{j}, x_{k})$ and posed the question whether instance $x_{i}$ was more similar to $x_{j}$ than to $x_{k}$. Qian \textit{et al.} \cite{Qian_2013} proposed an active learning algorithm where the query strategy was to place an ordering (or partial ordering) on the similarity of the neighbors of the selected unlabeled sample, rather than querying its actual label.

\textbf{Active Learning for Video Classification}: While AL has been extensively studied for image recognition \cite{Yoo_2019, Aditya_2019_WACV, Sinha_2019, Joshi_2012}, it is much less explored for video classification. Similar to image classification, uncertainty sampling (using metrics like entropy, error reduction) is a popular AL query strategy for video recognition \cite{ECML_Paper, Transportation_Journal}. Yan \textit{et al.} \cite{Yan_2003} proposed a multi-class AL framework for video classification using expected error reduction. Since the estimation of the posterior probability distribution $P(y|x)$ may be unreliable due to the lack of sufficient training data, simple heuristics were also proposed to simplify the sample selection strategies. Another approach was developed in the context of SVMs, which queried a set of samples which can produce the maximum expected reduction in the SVM objective \cite{SVM_AL_1, SVM_AL_2}. Bandla and Grauman \cite{Bandla_2013} used AL to train an action detector for videos which selected the video which was expected to maximally reduce the entropy among all unlabeled videos. The core idea was to use the current trained detector to extract relevant portions in the video where the action of interest occurs, so that the video segment outside the interval does not introduce noise in the entropy computation. However, this method is specifically designed to actively learn an action detector from videos. Active contrastive learning has also been explored for learning audio-visual representations from unlabeled videos \cite{Contrastive_AL}. 

All these methods require the human annotator to watch an unlabeled video end-to-end in order to provide a label, which may be extremely time-consuming and arduous. In contrast, our framework identifies a subset of exemplar frames, and the human labeler has to label a video by merely reviewing the frames, which is a much more efficient annotation strategy. Our method is applicable to any type of videos and does not make any assumptions about the contents of the video. Other related efforts include AL for video tracking \cite{AL_Video_Tracking}, video description \cite{AL_Video_Description}, video recommendation \cite{AL_Video_Recommendation} and video segmentation \cite{AL_Video_Segmentation}. However, these methods attempt to solve a different problem than video classification, which is the focus of this paper. We now describe our framework.

\section{Proposed Framework}
\label{sec_proposed}


Consider an active learning problem for video classification, where we are given a labeled training set $L$ and an unlabeled set $U$, with $|L| \ll |U|$. Each data sample $x$ in $L$ and $U$ is a video. Let $w$ be the deep neural network trained on $L$ and $C$ be the number of classes in the data. Our objective is two-fold: $(i)$ select a batch $B$ containing $b$ unlabeled videos so that the model trained on $L \cup B$ has maximum generalization capability; $(ii)$ however, we are not allowed to show an entire video to a human annotator and ask for its label; we are required to select a subset of $k$ exemplar frames from each queried video, so that only those can be shown to an annotator for labeling the video. 

Both these objectives are critical in improving the generalization capability of the deep model. The first objective ensures that the salient videos are selected from the unlabeled set for active query. The second objective ensures that the most representative frames are selected from each video for query. This is important, as otherwise, the annotator may not be confident enough to provide a label or may provide an incorrect label, both of which will result in a wastage of query budget and  degrade the performance of the model. In the following sections, we discuss our active sampling strategies for sampling videos and frames. 

\subsection{Active Video Sampling}
\label{subsec_AVS}

We quantified the utility of a batch of $b$ videos and selected a batch furnishing the maximal utility. The informativeness and diversity metrics were used to compute the utility of a batch of videos in this research. An active learning framework driven by these conditions ensures that the video samples in the batch augment useful knowledge to the underlying deep neural network, and there is high diversity (minimum redundancy) of information among the samples in the batch. These conditions have been used in previous active learning research \cite{Shen_2004}.

\textbf{Computing informativeness}: The informativeness of an unlabeled video sample $x_{i}$ was computed as the uncertainty of the deep model $w$ in predicting a label for $x_{i}$. The Shannon's entropy was used to compute the prediction uncertainty:

\begin{equation}
\label{eqn_info}
e(x_{i}) = -\sum_{y=1}^{C} P(y|x_{i}, w) \log P(y|x_{i}, w)
\end{equation} 

\textbf{Computing diversity}: We computed a diversity matrix $R \in \Re^{|U| \times |U|}$ where $R(i,j)$ denotes the diversity between videos $x_{i}$ and $x_{j}$ in the unlabeled set. We used the kernelized distance on the deep feature representations to compute the diversity between a pair of videos in this research:

\begin{equation}
\label{eqn_div}
R(i,j) = K (x_{i}, x_{j})
\end{equation} 

\noindent where $K  = (. , .)$ denotes the distance in the Reproducing Kernel Hilbert Space (RKHS) \cite{Sriperumbudur_2011}.

\subsubsection{Active Video Selection}
\label{subsubsec_ASS}

By definition, all the entries in $e$ and $R$ are non-negative, that is, $e_{i} \geq 0$ and $R(i,j) \geq 0, \forall i,j$. Given $e$ and $R$, our objective is to select a batch of videos with high uncertainties (given by the entries in $e$) and high mutual diversity (given by the entries in $R$). We define a binary selection vector $z \in \Re^{|U| \times 1}$ where $z_{i}$ denotes whether the unlabeled video $x_{i}$ will be selected in the batch $(z_{i} = 1)$ or not $(z_{i} = 0)$. Our batch selection task (with batch size $b$) can thus be posed as the following NP-hard integer quadratic programming (IQP) problem:

\begin{equation*}
\label{eqn_optimization_1}
\max_{z} e^{T} z + \mu z^{T} R z
\end{equation*}
\begin{equation}
\label{eqn_M_summation_new_1}
\text{s.t.} \hspace{0.1in} z_{i} \in \{0,1\}, \forall i \hspace{0.1in} \text{and} \hspace{0.1in} \sum_{i=1}^{|U|} z_{i} = b
\end{equation}

\noindent where $\mu$ is a weight parameter governing the relative importance of the two terms. The binary integer constraints on $z$ allow us to combine $e$ and $R$ into a single matrix $Q \in \Re^{|U| \times |U|}$ and express the optimization problem as follows:
\begin{equation*}
\label{eqn_optimization}
\max_{z} z^{T} Q z
\end{equation*}
\begin{equation}
\label{eqn_M_summation_new}
\text{s.t.} \hspace{0.1in} z_{i} \in \{0,1\}, \forall i \hspace{0.1in} \text{and} \hspace{0.1in} \sum_{i=1}^{|U|} z_{i} = b
\end{equation}

\noindent where the matrix $Q$ is constructed as follows:
\begin{equation}
\label{eqn_Q}
Q(i,j) = 
\begin{cases}
\mu R(i,j), & \text{if} \hspace{.03in} i\neq j\\
e(i),   & \text{if} \hspace{.03in} i = j
\end{cases}
\end{equation}

The binary integer constraints on the variable $z$ make the IQP in Equation (\ref{eqn_M_summation_new}) NP-hard. We used the \textit{Iterative Truncated Power} algorithm \cite{Yuan_JMLR_2013} to solve this optimization problem. 

\subsubsection{The Iterative Truncated Power Algorithm}

This algorithm was originally proposed in the context of the sparse eigenvalue and the densest $k$-subgraph problems. It attempts to solve an optimization problem similar to that in Equation (\ref{eqn_M_summation_new}). The algorithm starts with an initial solution $z_{0}$ and then generates a sequence of solutions $z_{1}, z_{2}, \ldots $. The solution $z_{t}$ at iteration $t$ is obtained by multiplying the solution $z_{t-1}$ at iteration $(t-1)$ by the matrix $Q$ and then truncating all the entries to $0$, except the $b$ largest entries. The process is repeated until convergence. The algorithm is guaranteed to converge monotonically for a positive semi-definite (psd) matrix $Q$. When the matrix $Q$ is not psd, the algorithm can be run on the shifted quadratic function (with a positive scalar added to the diagonal elements) to guarantee a monotonic convergence \cite{Yuan_JMLR_2013}. The algorithm is computationally efficient and converges fast. It benefits from a good starting point. In our empirical studies, the initial solution $z_{0}$ was taken as the indicator vector corresponding to the $b$ largest column sums of the matrix $Q$, as it produced competitive results in our preliminary experiments. The pseudo-code for our active video sampling algorithm is presented in Algorithm \ref{alg_AVS}.

\begin{algorithm}
\caption{The Proposed Active Video Sampling Algorithm}
\label{alg_AVS}
\begin{algorithmic}[1]
\REQUIRE{Training set $L$, Unlabeled set $U$, batch size $b$ and weight parameter $\mu$}\newline
\STATE Train a deep neural network $w$ on the training set $L$
\STATE Compute the entropy vector $e$ (Equation (\ref{eqn_info})) and the diversity matrix $R$ (Equation (\ref{eqn_div})) 
\STATE Compute the matrix $Q$, as described in Equation (\ref{eqn_Q})
\STATE Derive the initial solution $z_{0}$ as the indicator vector containing the $b$ largest column sums of the matrix $Q$
\STATE t = 1
\REPEAT
\STATE Compute $\widehat{z_{t}} = Q.z_{t-1}$
\STATE Identify $F_{t}$ as the index set of $\widehat{z_{t}}$ with top $b$ values
\STATE Set $z_{t}$ to be $1$ on the index set $F_{t}$ and $0$ otherwise
\STATE t = t + 1
\UNTIL Convergence
\STATE Select a batch of $b$ unlabeled videos based on the final solution $z_{t}$
\end{algorithmic}
\end{algorithm}

\subsubsection{Computational Considerations}

Computing the diversity matrix $R$ involves quadratic complexity. We first note that $R$ needs to be computed only once in our framework, before the start of the AL iterations. As the unlabeled videos get queried through AL, we can keep deleting the corresponding rows and columns in $R$ to derive the new diversity matrix. Moreover, random projection algorithms can be used to speed up computations. The theory of random projections states that, if we have a point cloud in a high dimensional space, they may be projected into a suitable lower-dimensional space such that the distances between the points are approximately preserved \cite{Random_Proj}. A data matrix $A \in \Re^{N \times D}$ in the $D$ dimensional space is multiplied by a random projection matrix $X \in \Re^{D \times d}$ $(d \ll D)$ to obtain a projected matrix $B \in \Re^{N \times d}$ in the lower dimensional space $d$: $B = AX$ \cite{Vempala_2004}. This can be used to substantially reduce the computational overhead, as distance computations are more efficient in the low dimensional space. We will explore this as part of future research.

\subsection{Active Frame Sampling}
\label{subsec_AFS} 

Once we select $b$ videos from the unlabeled set, our next task is to identify a subset of $k$ frames from each of these videos; we exploited representative sampling techniques for this purpose. These techniques identify the exemplar data points which well-represent a given dataset. In particular, the \textit{coreset} algorithm selects a subset of points such that a model trained over the selected subset is maximally similar to that trained on the whole dataset. For the sake of completeness, we discuss the main ideas here and request interested readers to refer to \cite{coreset_AL} for further details. Coreset poses the subset selection problem as:

\begin{equation}
\min_{s: |s|=k} \Bigg | \frac{1}{n}\sum_{i \in [n]} l(x_{i}, y_{i}, A_{i}) - \frac{1}{|s|} \sum_{j \in s} l(x_{j}, y_{j}, A_{j}) \Bigg |
\end{equation}

\noindent where $(x_{i}, y_{i})$ denotes a training sample and its label, $A_{i}$ denotes a learning algorithm which outputs a set of parameters by minimizing a loss function $l(. , . , .)$ on a given labeled set $i$. Informally, given a budget $k$, the goal is to select a set of samples $s$, such that the model trained on $s$ depicts similar performance as the model trained on the whole dataset with $n$ samples. 

This function cannot be directly optimized, as the labels of the samples in the unlabeled set are unknown. An upper bound of this function was derived and the problem of active sampling was shown to be equivalent to the $k$-center problem (also called min-max facility location problem) \cite{Reza_2009}. The objective of this problem is to select $k$ center points from $n$ samples, such that the largest distance between a data point and its nearest center is minimized. Formally, this can be posed as follows: 

\begin{equation}
\min_{s: |s| = k} \max_{i} \min_{j \in s} \Delta(x_{i}, x_{j})
\end{equation}

This problem is NP-Hard \cite{Cook_1998}. However, a greedy algorithm, as detailed in Algorithm \ref{alg_AFS}, is guaranteed to produce a solution $s$ such that: $\max_{i} \min_{j \in s} \Delta(x_{i}, x_{j}) \leq 2 \times OPT$, where $OPT$ is the optimal solution. We used this algorithm to select a subset of $k$ frames from each of the queried videos. As evident from the formulation, our method does not make any assumptions about the contents of the video, and is applicable to any type of video. 

\begin{algorithm}
\caption{The Active Frame Sampling Algorithm \cite{coreset_AL}}
\label{alg_AFS}
\begin{algorithmic}[1]
\REQUIRE{A video with $n$ frames ($x_{1}, x_{2}, \ldots x_{n}$) and frame budget $k$}\newline
\STATE Initialize $s = \Phi$
\REPEAT
\STATE $u = \argmax_{i \in [n] \backslash s} \min_{j \in s} \Delta (x_{i}, x_{j})$
\STATE $s = s \cup \{u\}$
\UNTIL $|s| = k$
\STATE Select $k$ frames from the video based on the set $s$
\end{algorithmic}
\end{algorithm}


\section{Experiments and Results}
\label{sec_expt}


\subsection{Datasets}

We used the \textbf{UCF-101} \cite{Soomro_TR2012} and the \textbf{Kinetics} datasets \cite{Kinetics_dataset} to study the performance of our algorithm. Both these datasets contain videos of humans performing a variety of actions, captured under unconstrained, real-world conditions, and are extensively used to study the performance of video classification algorithms. We used data from $5$ classes at random from each dataset for our experiments. 

\subsection{Oracle Simulation} 

All the publicly available video datasets contain annotations for the complete videos; we did not find any datasets which contain annotations based on a subset of frames. Also, different active sampling algorithms will select different subsets of frames, and it is challenging to obtain annotations from a human labeler for every possible subset of frames for a given video, to conduct experiments. We therefore used a deep neural network to simulate the human labeling oracle in our empirical studies. The oracle model was trained on a completely different subset of the data. No information about the oracle model was used in the design and development of our active learning algorithm. During AL, when a video sample was selected for query, the selected frames were passed as an input to the trained oracle model and its prediction entropy on the sample was computed. If the entropy exceeded a particular threshold $\tau_{oracle}$, the oracle was assumed to be not confident enough to produce a label, and no label was returned; otherwise, the oracle returned the predicted label (which may be correct or incorrect). These were done to appropriately mimic a real-world data annotation setup with a human annotator. 

\subsection{Implementation Details}

\textbf{Base Model}: We used a CNN-RNN architecture in our experiments where InceptionV3 pretrained on the ImageNet-1k dataset was used as the feature extractor and a GRU network as the decoder \footnote{\url{https://keras.io/examples/vision/video_classification/}}. The input frames were scaled and normalized to a fixed input size of $224 \times 224$ pixels and fed into the Convolutional Neural Network (CNN). The features extracted were fed into a $5$-layer GRU network which consists of $2$ GRU layers and $1$ fully connected layer with one dropout layer. The $2$ GRU layers had $20$ and $12$ neurons, while the first fully connected layer had $8$ neurons with the \textit{ReLU} activation function. 
We used the \textit{adam} optimizer with a learning rate of $0.001$, momentum of $0.99$, batch size of $32$, and the network was trained for $20$ epochs in each active learning iteration.

\textbf{Oracle Model}: We used a similar CNN-RNN architecture as the oracle model. However, for the oracle model, the $2$ GRU layers of the GRU network had $40$ and $16$ neurons. We used the \textit{adam} optimizer with a learning rate of $0.001$ for the UCF dataset and $0.01$ for the Kinetics dataset, momentum of $0.99$, batch size of $64$, and the network was trained for $30$ epochs. As part of future research, we plan to study the performance of our framework with other architectures for the oracle model, and also conduct experiments with real people as annotators.

\subsection{Experimental Setup}

Each dataset was split into $5$ parts: $(i)$ an initial training set $L$; $(ii)$ unlabeled set $U$; $(iii)$ test set $T$; $(iv)$ training set to train the oracle model $L_{oracle}$; and $(v)$ test set $T_{oracle}$ to test the oracle model and compute the entropy threshold $\tau_{oracle}$. The number of samples (videos) in each of these sets, together with the accuracy of the oracle model ($A_{oracle}$) for each dataset are depicted in Table \ref{tab_dataset_split}. We note that a better trained oracle could have potentially improved the performance of our algorithm; however, we wanted to validate our algorithm in a challenging real-world setup, where the annotators can abstain from labeling samples and can also provide incorrect labels. We therefore used an oracle model with moderate accuracy ($ \approx 70 - 75\%$) in our empirical studies.

\begin{table}[h]
	\centering
	\normalsize
	
	\begin{tabular}{|c|c|c|c|c|c|c|}
      \hline
   
      Dataset & L & U & T & \textbf{$L_{oracle}$} & \textbf{$T_{oracle}$} & \textbf{$A_{oracle}$} \\
       \hline
       UCF & 250  & 320 & 150 & 697 & 185 & $75.61\%$\\
       \hline
       Kinetics & 500 & 750 & 300 & 584 & 211 & $71.34\%$\\
       \hline
      
       \end{tabular}

\caption{Dataset split used in our experiments, together with the accuracy of the oracle model for the two datasets.}
	\label{tab_dataset_split}
\end{table}

\begin{figure*}[ht]
	\centering
		\subfigure[UCF]{
          \label{fig_ucf}
          \includegraphics[trim = 1.3in 3.2in 1.7in 3.2in,clip,width=.33\textwidth]{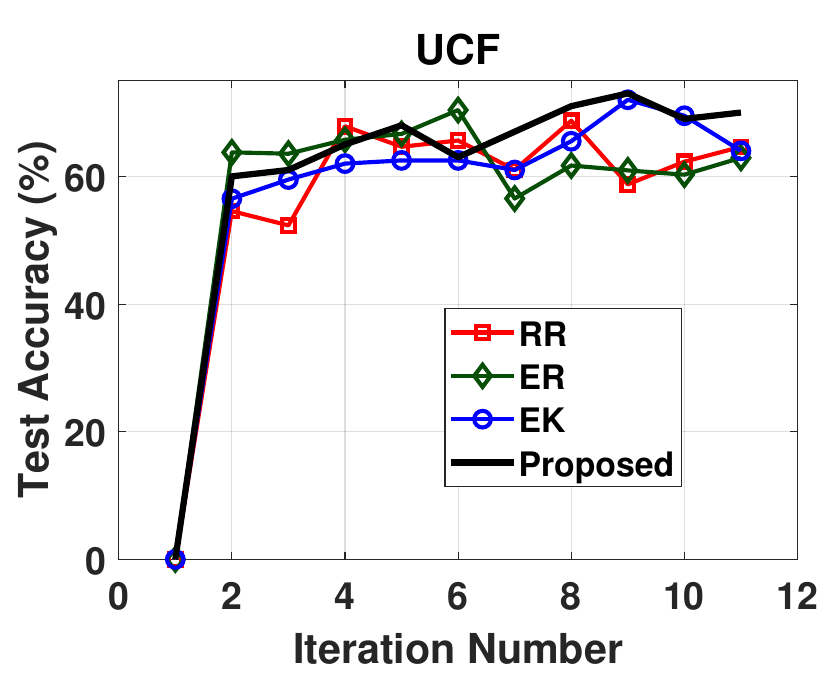}}   
     \hspace{.03in}
     \subfigure[Kinetics]{
          \label{fig_kinetics}
          \includegraphics[trim = 1.3in 3.2in 1.7in 3.2in,clip,width=.33\textwidth]{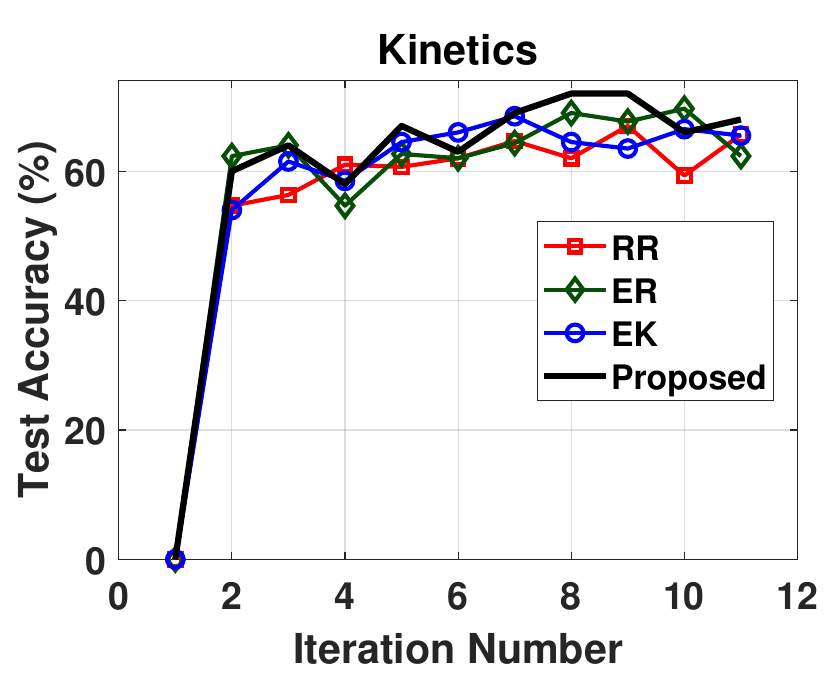}} 
      
      \caption{AL performance comparison. The $x$-axis denotes the iteration number and the $y$-axis denotes the accuracy on the test set. Best viewed in color.}
	\label{fig_main_results}
\end{figure*}

\begin{table*}[ht]
\centering
	\normalsize
    \begin{tabular}{|c|c|c|c|c|}
    \hline 
        \multicolumn{5}{|c|}{UCF Dataset: Video budget $b = 25$, Frame Budget $k=100$}   \\
    \hline
        Methods &  Total Videos Queried & Correct (\%) & Incorrect (\%) & Discarded (\%) \\
    \hline
        RR & 250 & $54.4 \pm 1.38$ & $42.4 \pm 1.6$ & $3.2 \pm 0.8$ \\
    \hline
        ER & 250 & $55.2 \pm 1.05$ & $40.4 \pm 1.05$ & $4.4 \pm 0.69$ \\
    \hline
        EK & 250 & $53.06 \pm 1.40$ & $30.93 \pm 2.20$ & $16 \pm 0.8$  \\
    \hline
        Proposed & 250 & $58.66 \pm 1.8$ & $31.06 \pm 0.61$ & $10.26 \pm 2.41$  \\
    \hline
    \end{tabular}
    
    \caption{Performance of the oracle on the UCF dataset.}
	\label{tab_oracle_UCF}
\end{table*}

\begin{table*}[ht]

\centering
	\normalsize
    \begin{tabular}{|c|c|c|c|c|}
    \hline 
        \multicolumn{5}{|c|}{Kinetics Dataset: Video budget $b = 25$, Frame Budget $k=100$}   \\
    \hline
        Methods &  Total Videos Queried & Correct (\%) & Incorrect (\%) & Discarded (\%) \\
    \hline
        RR & 250 & $67.33 \pm 1.40$ & $23.33 \pm 1.40$ & $9.33 \pm 1.40$ \\
    \hline
        ER & 250 & $64 \pm 1.2$ & $23.86 \pm 2.83$ & $12.13 \pm 4.02$ \\
    \hline
        EK & 250 & $64.66 \pm 1.28$ & $25.06 \pm 3.02$ & $10.26 \pm 4.31$  \\
    \hline
        Proposed & 250 & $66 \pm 1.44$ & $22.93 \pm 1.51$ & $11.06 \pm 1.28$  \\
    \hline
    \end{tabular}
        
    \caption{Performance of the oracle on the Kinetics dataset.}
	\label{tab_oracle_Kinetics}
\end{table*}

The oracle model was trained on $L_{oracle}$; each sample in $T_{oracle}$ was then passed as an input to the trained oracle and the prediction entropy was noted. The $50^{th}$ percentile of the prediction entropy distribution was taken as the entropy threshold $\tau_{oracle}$; during the AL iterations, if the entropy of any queried video exceeded this threshold, the oracle was assumed to abstain from labeling. 
The base model was first trained on the set $L$. In each AL iteration, each algorithm queried $b$ videos from the set $U$, and $k$ frames from each of the $b$ videos. The $k$ frames of each video were then passed as an input to the oracle model. Based on its prediction entropy on the sample, the oracle may or may not furnish a label for a given unlabeled video sample. If the oracle does not furnish a label for a given video, it was discarded. The other unlabeled videos (which were labeled by the oracle), together with the returned labels were then appended to the training set, the base model was updated, and its accuracy was computed on the test set. The process was repeated for $10$ iterations, which was taken as the stopping criterion in this work. All the results were averaged over $3$ runs (with different training, unlabeled and test sets) to rule out the effects of randomness. The video budget $b$ was taken as $25$ and the frame budget $k$ as $100$ in each AL iteration. The weight parameter $\mu$ in Equation (\ref{eqn_M_summation_new_1}) was taken as $0.01$ and a Gaussian kernel was used to compute the diversity matrix in Equation (\ref{eqn_div}).

%

\subsection{Comparison Baselines}

As mentioned in Section \ref{sec_related}, existing AL techniques for video classification query the complete videos for annotation and the labels obtained are assumed to be always correct. In our framework, the labeling oracle may refuse to provide a label to a queried video and may also provide an incorrect label. This is a more challenging and real-world scenario. It will thus not be fair to compare our method against the existing techniques. We used three comparison baselines in this work: $(i)$ \textbf{Random-Random (RR)}, where we selected a batch of $b$ videos at random and a subset of $k$ frames from each video at random; $(ii)$ \textbf{Entropy-Random (ER)}, where the $b$ videos with the highest classification entropies were queried and $k$ frames were queried from each at random; and $(iii)$ \textbf{Entropy-kmeans (EK)}, where $b$ videos were first selected using entropy sampling; $k$-means clustering was then performed and the $k$ frames corresponding to the $k$ cluster centroids were selected for query from each video. 
 
%
%
%

\begin{figure*}[ht]
	\centering
		\subfigure[Frame Budget k = 10]{
          \label{fig_FS_10}
          \includegraphics[trim = 1.3in 3.2in 1.7in 3.2in,clip,width=.23\textwidth]{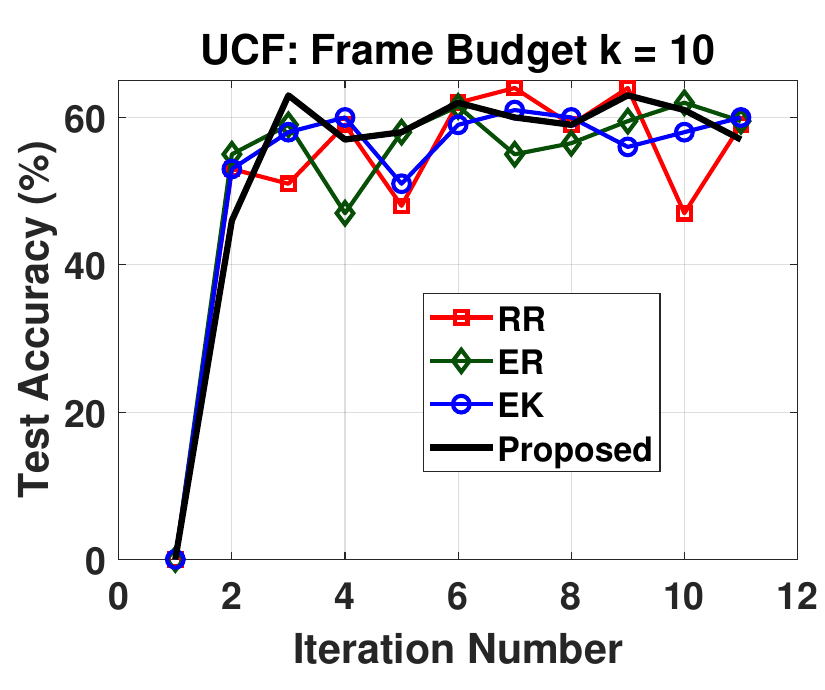}}   
     \hspace{.03in}
     \subfigure[Frame Budget k = 20]{
          \label{fig_FS_20}
          \includegraphics[trim = 1.3in 3.2in 1.7in 3.2in,clip,width=.23\textwidth]{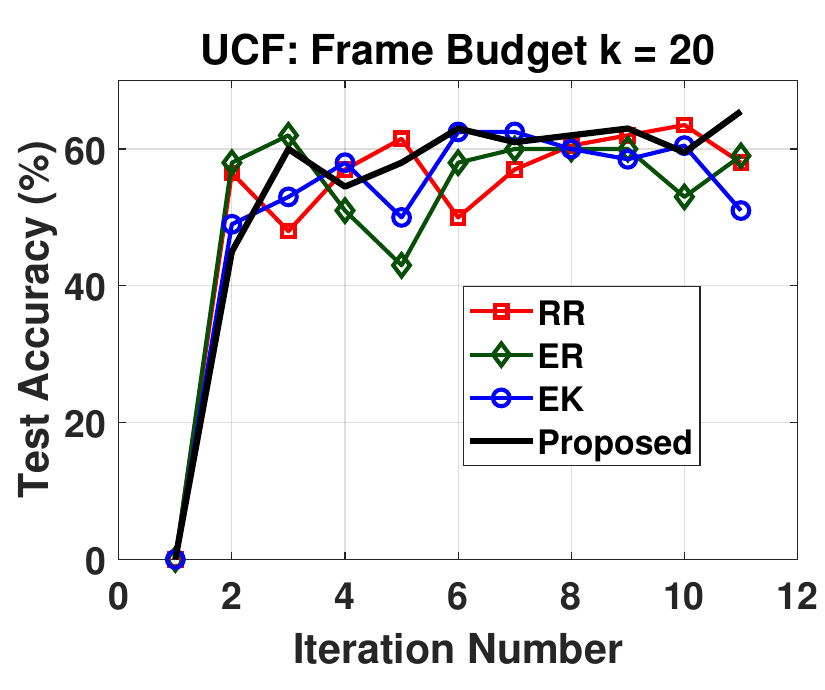}}
     \hspace{.03in}
		 \subfigure[Frame Budget k = 50]{
          \label{fig_FS_50}
          \includegraphics[trim = 1.3in 3.2in 1.7in 3.0in,clip,width=.23\textwidth]{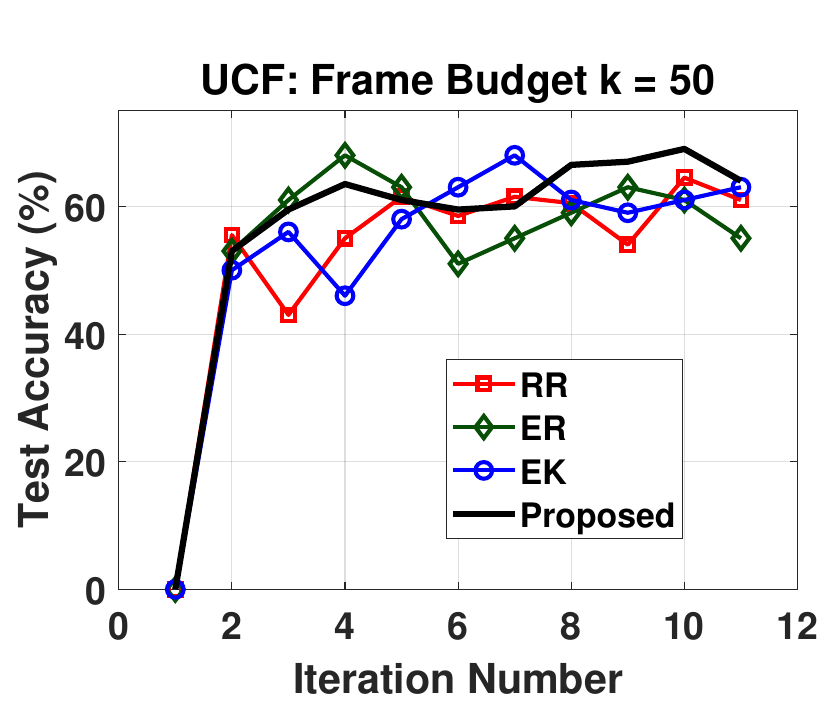}}        
      \hspace{.03in}
		 \subfigure[Frame Budget k = 100]{
          \label{fig_FS_100}
          \includegraphics[trim = 1.3in 3.2in 1.7in 3.0in,clip,width=.23\textwidth]{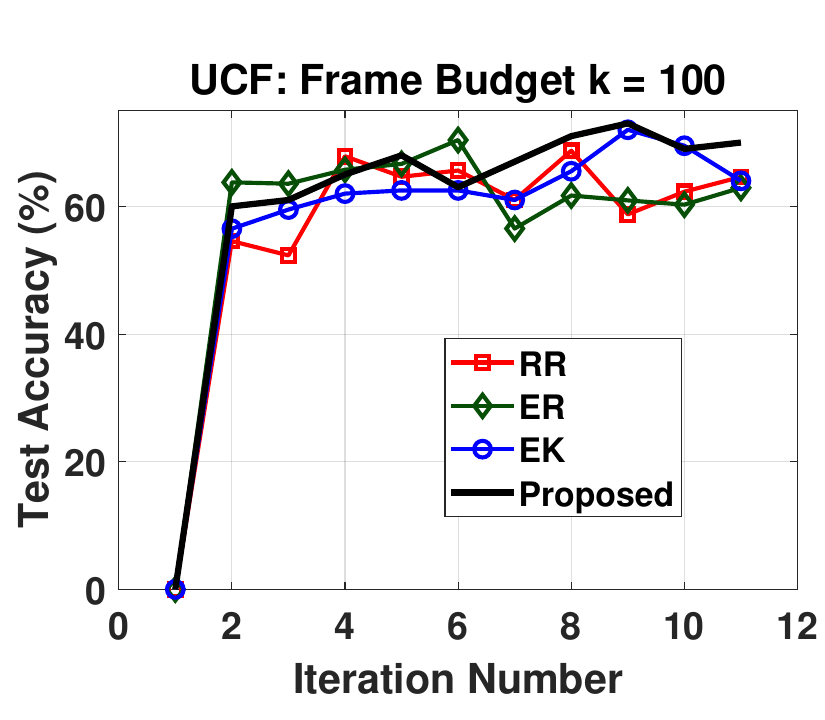}}        
      
      \caption{Study of the number of queried frames per video. The result with $k = 100$ (default setting) is the same as in Figure \ref{fig_ucf} and is included here for the sake of completeness. Best viewed in color.}
	\label{fig_FS_results}
\end{figure*}

\begin{figure*}[ht]
	\centering
		\subfigure[Video Budget b = 15]{
          \label{fig_BS_15}
          \includegraphics[trim = 1.3in 3.2in 1.7in 3.2in,clip,width=.23\textwidth]{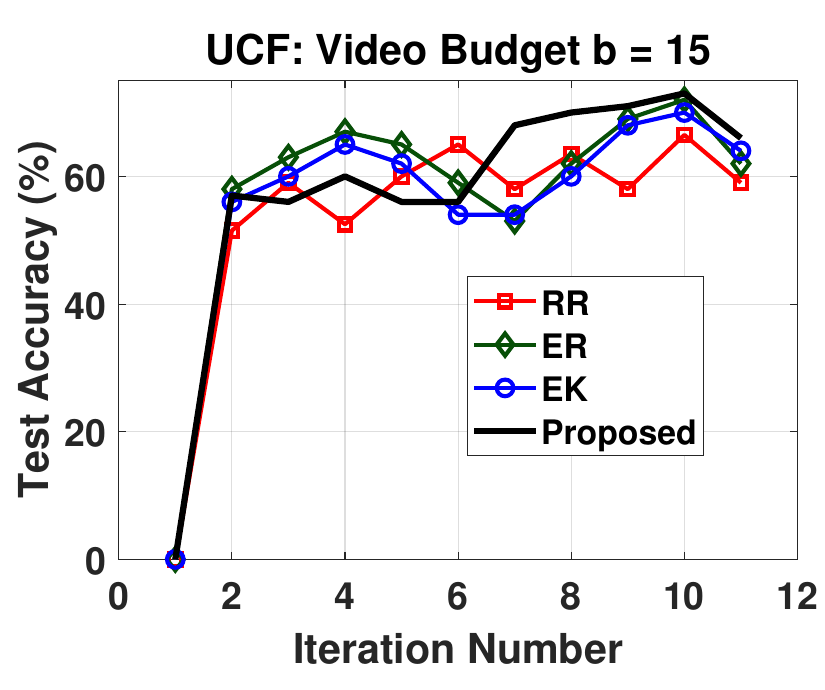}}   
     \hspace{.03in}
     \subfigure[Video Budget b = 20]{
          \label{fig_BS_20}
          \includegraphics[trim = 1.3in 3.2in 1.7in 3.2in,clip,width=.23\textwidth]{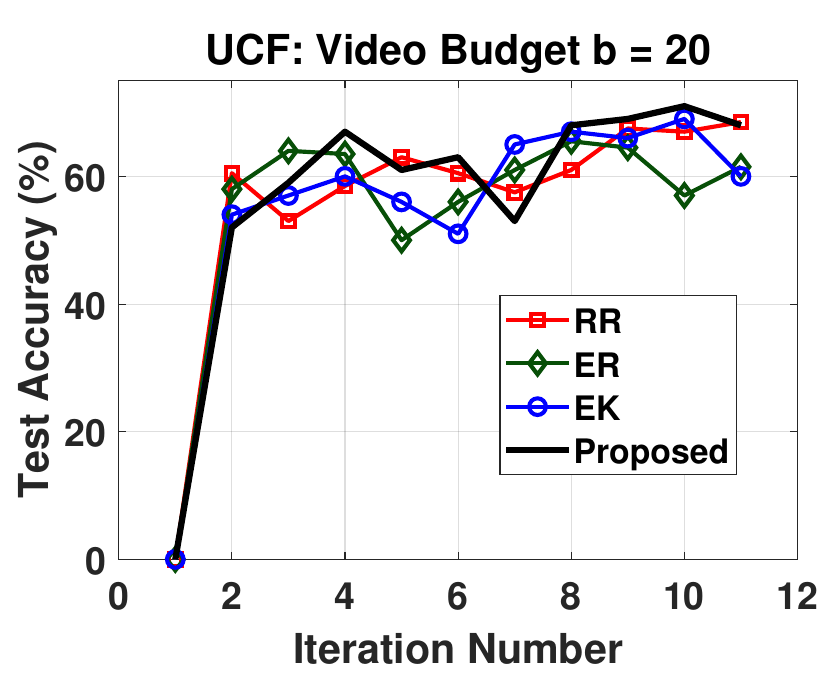}}
     \hspace{.03in}
		 \subfigure[Video Budget b = 25]{
          \label{fig_BS_25}
          \includegraphics[trim = 1.3in 3.2in 1.7in 3.0in,clip,width=.23\textwidth]{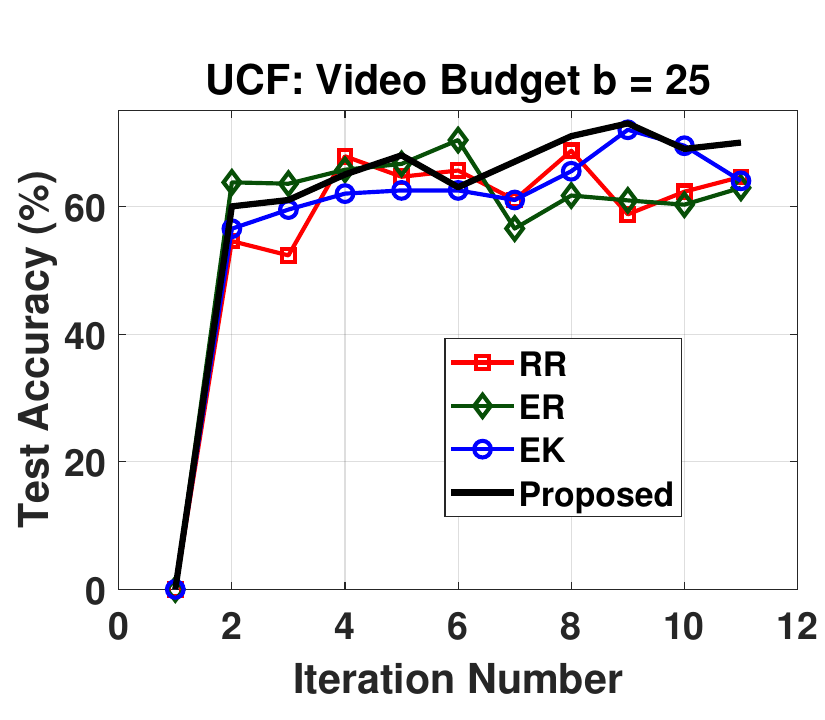}}        
      \hspace{.03in}
		 \subfigure[Video Budget b = 30]{
          \label{fig_BS_30}
          \includegraphics[trim = 1.3in 3.2in 1.7in 3.0in,clip,width=.23\textwidth]{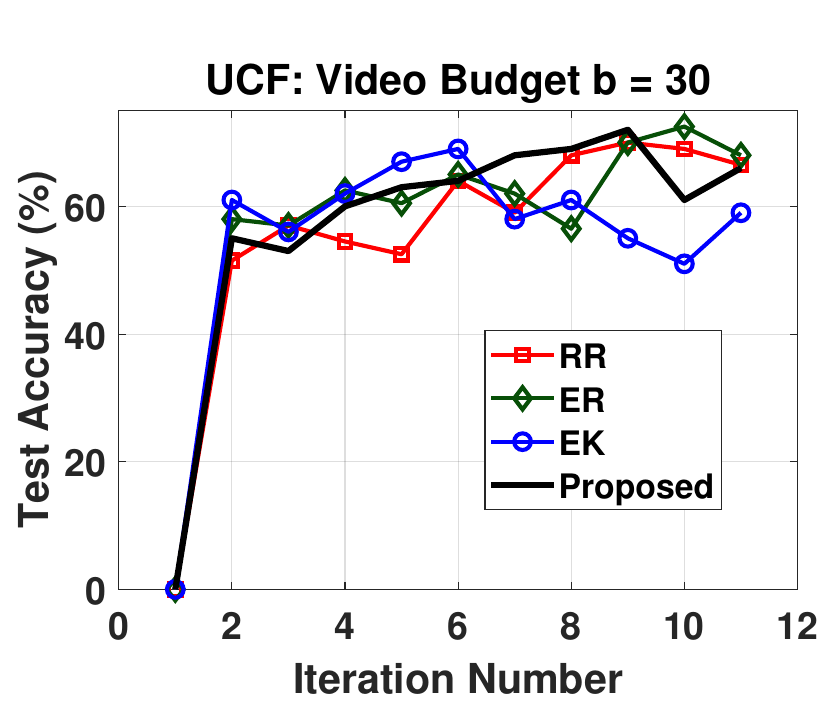}}        
      
      \caption{Study of the number of queried videos in each AL iteration. The result with $b = 25$ (default setting) is the same as in Figure \ref{fig_ucf} and is included here for the sake of completeness. Best viewed in color.}
	\label{fig_BS_results}
\end{figure*}

\subsection{Active Learning Performance}

The AL performance results are depicted in Figure \ref{fig_main_results}. In each figure, the $x$-axis represents the iteration number, and the $y$-axis denotes the accuracy on the test set. The proposed method comprehensively outperforms the \textit{RR} method on both datasets. The \textit{ER} method depicts random fluctuations in the test accuracy over the AL iterations; our method, on the other hand, depicts a more steady growth in the test accuracy. The \textit{EK} method depicts the best performance among the baselines, but is not as good as our method. Our method outperforms \textit{EK} in most of the AL iterations across both the datasets. It also attains the highest accuracy after $10$ AL iterations for both the datasets. We can conclude the following: $(i)$ our video selection criterion based on uncertainty and diversity identifies the most informative videos in the unlabeled set; and $(ii)$ our frame selection criterion based on representative sampling selects a subset of exemplar frames from each queried video, so that a large percentage of them can be correctly annotated by the oracle, which enriches the quality of our training data. As a result, our method augments maximal useful information to the deep neural network, which boosts its generalization capability. These results unanimously corroborate the potential of our framework in substantially reducing the human annotation effort in real-world video classification applications, where labeling a single sample involves significant time and human labor. 

The performance of the oracle model is reported in Tables \ref{tab_oracle_UCF} and \ref{tab_oracle_Kinetics} for the UCF and Kinetics datasets respectively. A total of $250$ videos were queried from these datasets ($25$ videos in each of the $10$ AL iterations). The tables show the percentage of these videos that were correctly annotated, incorrectly annotated and discarded by the labeling oracle. For the UCF dataset, and for the proposed method, the oracle correctly annotated $58.66\%$ of the queried videos (the highest among all the methods). This shows that representative sampling through coreset is an effective strategy to identify the exemplar frames from a queried video, which have a high probability of receiving the correct label from the oracle, and augmenting useful information to the training set. For the Kinetics dataset, $66\%$ of the videos queried by our method were correctly annotated by the oracle, where as $67.33\%$ of the videos queried by \textit{Random Sampling} were annotated correctly by the oracle. However, we note that, having a high percentage of unlabeled videos correctly labeled by the oracle does not necessarily mean that the useful samples are being queried. For instance, it is easy to select a batch of videos, which do not have much useful content and are easy to label, and get a high percentage of them correctly labeled by the oracle. However, these videos, even if correctly labeled, will not augment much useful information to the training set, as they are devoid of any useful content. Even though \textit{RR} depicts a slightly higher percentage of correctly labeled samples than our method in Table \ref{tab_oracle_Kinetics}, its generalization accuracy is much worse than our method, as evident from Figure \ref{fig_kinetics}. The key challenge is to \textit{query a set of informative videos \textbf{and} get a high percentage of them correctly labeled by the oracle}; both of these are crucial in improving the generalization capability of the model over time. The results in Figure \ref{fig_main_results} jointly capture both these aspects, and show that our method outperforms the baselines.

\subsection{Effect of the Number of Queried Frames per Video}

In this experiment, we studied the effect of the frame budget $k$ (number of frames allowed to be selected from a queried video) on the AL performance. The results on the UCF dataset, with frame budgets $10$, $20$, $50$ and $100$ are presented in Figure \ref{fig_FS_results}. Our method depicts impressive performance across different frame budgets. For frame budgets $20$, $50$ and $100$, our framework attains the highest test accuracy after $10$ AL iterations. Note that querying lesser number of frames from a video lessens the labeling burden on the oracle, as the oracle has to review an even smaller number of frames to furnish a label. These results show the promise and potential of our technique to further reduce human annotation effort in a video classification application. 

\subsection{Effect of the Number of Queried Videos}

The goal of this experiment was to study the effect of the video budget $b$ (number of videos queried in each AL iteration) on the AL performance. The results on the UCF dataset with $b = 15, 20, 25$ and $30$ are shown in Figure \ref{fig_BS_results}. Our framework once again surpasses the baselines across the different budgets. These results are important from the standpoint of a real-world application, where the batch size is governed by the time, man-power and other available resources in a given application, and is different for different applications.

\section{Conclusion and Future Work}
\label{sec_conc}

The goal of this research was to devise an efficient annotation mechanism to reduce human annotation effort in video classification applications, where annotating a single data instance is extremely tedious. Our framework identifies a batch of informative videos, together with a set of exemplar frames from each; the human annotator has to produce a label for each video just by reviewing the subset of frames, instead of watching the complete video end-to-end. To the best of our knowledge, this is the first research effort to develop an AL technique for video classification, with this kind of a query and annotation mechanism. Our empirical results validated the promise and potential of our framework to drastically reduce human annotation effort in training a deep neural network for video classification. We hope this research will motivate the development of AL algorithms with other annotation mechanisms, with the goal of further reducing the human annotation effort in video classification. 

As part of future work, we plan to validate the performance of our algorithm on other applications where the data has a temporal nature. For instance, the proposed query mechanism will also be very relevant in a text classification application to identify informative text snippets, so that a human annotator can furnish a label by reviewing only the snippets, rather than reading the document end-to-end. We will also study the performance of our framework with different size of the data splits, as outlined in Table \ref{tab_dataset_split}.

\bibliographystyle{IEEEtran}
\bibliography{IJCNN_2023_bib}

\begin{thebibliography}{10}
\providecommand{\url}[1]{#1}
\csname url@samestyle\endcsname
\providecommand{\newblock}{\relax}
\providecommand{\bibinfo}[2]{#2}
\providecommand{\BIBentrySTDinterwordspacing}{\spaceskip=0pt\relax}
\providecommand{\BIBentryALTinterwordstretchfactor}{4}
\providecommand{\BIBentryALTinterwordspacing}{\spaceskip=\fontdimen2\font plus
\BIBentryALTinterwordstretchfactor\fontdimen3\font minus
  \fontdimen4\font\relax}
\providecommand{\BIBforeignlanguage}[2]{{%
\expandafter\ifx\csname l@#1\endcsname\relax
\typeout{** WARNING: IEEEtran.bst: No hyphenation pattern has been}%
\typeout{** loaded for the language `#1'. Using the pattern for}%
\typeout{** the default language instead.}%
\else
\language=\csname l@#1\endcsname
\fi
#2}}
\providecommand{\BIBdecl}{\relax}
\BIBdecl

\bibitem{Review_Paper}
V.~Sharma, M.~Gupta, A.~Kumar, and D.~Mishra, ``Video processing using deep
  learning techniques: A systematic literature review,'' \emph{IEEE Access},
  vol.~9, 2021.

\bibitem{Ng_2015}
J.~Ng, M.~Hausknecht, S.~Vijayanarasimhan, O.~Vinyals, R.~Monga, and
  G.~Toderici, ``Beyond short snippets: Deep networks for video
  classification,'' in \emph{IEEE Conference on Computer Vision and Pattern
  Recognition (CVPR)}, 2015.

\bibitem{Karpathy_Paper}
A.~Karpathy, G.~Toderici, S.~Shetty, T.~Leung, R.~Sukthankar, and L.~Fei-Fei,
  ``Large-scale video classification with convolutional neural networks,'' in
  \emph{IEEE Conference on Computer Vision and Pattern Recognition (CVPR)},
  2014.

\bibitem{Tian_2019}
H.~Tian, Y.~Tao, S.~Pouyanfar, S.~Chen, and M.~Shyu, ``Multimodal deep
  representation learning for video classification,'' \emph{World Wide Web},
  vol.~22, no.~3, pp. 1325 -- 1341, 2019.

\bibitem{Settles_2010}
B.~Settles, ``Active learning literature survey,'' in \emph{Technical Report:
  University of Wisconsin-Madison}, 2010.

\bibitem{Yoo_2019}
D.~Yoo and I.~Kweon, ``Learning loss for active learning,'' in \emph{IEEE
  Conference on Computer Vision and Pattern Recognition (CVPR)}, 2019.

\bibitem{tong_support_2000}
S.~Tong and D.~Koller, ``Support vector machine active learning with
  applications to text classification,'' \emph{Journal of Machine Learning
  Research (JMLR)}, vol.~2, pp. 45--66, 2001.

\bibitem{Hatice_2010}
H.~Osmanbeyoglu, J.~Wehner, J.~Carbonell, and M.~Ganapathiraju, ``Active
  machine learning for transmembrane helix prediction,'' \emph{BMC
  Bioinformatics}, vol.~11, no.~1, 2010.

\bibitem{Gorriz_2017}
M.~Gorriz, A.~Carlier, E.~Faure, and X.~G. i~Nieto, ``Cost-effective active
  learning for melanoma segmentation,'' in \emph{Neural Information processing
  Systems (NeurIPS) Workshop}, 2017.

\bibitem{Ren_2021}
P.~Ren, Y.~Xiao, X.~Chang, P.~Huang, Z.~Li, B.~Gupta, X.~Chen, and X.~Wang, ``A
  survey of deep active learning,'' \emph{ACM Computing Surveys}, vol.~54,
  no.~9, 2021.

\bibitem{Holub_2008}
A.~Holub, P.~Perona, and M.~Burl, ``Entropy-based active learning for object
  recognition,'' in \emph{IEEE Conference on Computer Vision and Pattern
  Recognition Workshops (CVPR-W)}, 2008.

\bibitem{freund_selective_1997}
Y.~Freund, S.~Seung, E.~Shamir, and N.~Tishby, ``Selective sampling using the
  query by committee algorithm,'' \emph{Machine Learning}, vol.~28, no. 2-3,
  pp. 133--168, 1997.

\bibitem{Fu_KDD_2018}
W.~Fu, M.~Wang, S.~Hao, and X.~Wu, ``Scalable active learning by approximated
  error reduction,'' in \emph{ACM SIGKDD International Conference on Knowledge
  Discovery and Data Mining}, 2018.

\bibitem{AL_Submod_1}
K.~Wei, R.~Iyer, and J.~Bilmes, ``Submodularity in data subset selection and
  active learning,'' in \emph{International Conference on Machine Learning
  (ICML)}, 2015.

\bibitem{AL_Submod_2}
K.~Fujii and H.~Kashima, ``Budgeted stream-based active learning via adaptive
  submodular maximization,'' in \emph{Neural Information Processing Systems
  (NeurIPS)}, 2016.

\bibitem{coreset_AL}
O.~Sener and S.~Savarese, ``Active learning for convolutional neural networks:
  A core-set approach,'' in \emph{International Conference on Learning
  Representations (ICLR)}, 2018.

\bibitem{Geifman_2019}
Y.~Geifman and R.~El-Yaniv, ``Deep active learning with a neural architecture
  search,'' in \emph{Neural Information Processing Systems (NeurIPS)}, 2019.

\bibitem{Shui_2020}
C.~Shui, F.~Zhou, C.~Gagne, and B.~Wang, ``Deep active learning: Unified and
  principled method for query and training,'' in \emph{International Conference
  on Artificial Intelligence and Statistics (AISTATS)}, 2020.

\bibitem{Ducoffe_2018}
M.~Ducoffe and F.~Precioso, ``Adversarial active learning for deep networks: a
  margin based approach,'' in \emph{International Conference on Machine
  Learning (ICML)}, 2018.

\bibitem{Mayer_2020}
C.~Mayer and R.~Timofte, ``Adversarial sampling for active learning,'' in
  \emph{IEEE Winter Conference on Applications of Computer Vision (WACV)},
  2020.

\bibitem{Zhang_2020}
B.~Zhang, L.~Li, S.~Yang, S.~Wang, Z.~Zha, and Q.~Huang, ``State-relabeling
  adversarial active learning,'' in \emph{IEEE Conference on Computer Vision
  and Pattern Recognition (CVPR)}, 2020.

\bibitem{Sinha_2019}
S.~Sinha, S.~Ebrahimi, and T.~Darrell, ``Variational adversarial active
  learning,'' in \emph{IEEE International Conference on Computer Vision
  (ICCV)}, 2019.

\bibitem{Chattopadhyay_2013}
R.~Chattopadhyay, W.~Fan, I.~Davidson, S.~Panchanathan, and J.~Ye, ``Joint
  transfer and batch-mode active learning,'' in \emph{International Conference
  on Machine Learning (ICML)}, 2013.

\bibitem{AL_RL}
D.~Krueger, J.~Leike, O.~Evans, and J.~Salvatier, ``Active reinforcement
  learning: Observing rewards at a cost,'' in \emph{Neural Information
  Processing Systems (NeurIPS) Workshop}, 2016.

\bibitem{Ruchansky_2015}
N.~Ruchansky, M.~Crovella, and E.~Terzi, ``Matrix completion with queries,'' in
  \emph{ACM Conference on Knowledge Discovery and Data Mining (KDD)}, 2015.

\bibitem{Molino_2017}
A.~Molino, X.~Boix, J.~Lim, and A.~Tan, ``Active video summarization:
  Customized summaries via on-line interaction with the user,'' in
  \emph{Association for the Advancement of Artificial Intelligence (AAAI)},
  2017.

\bibitem{Shim_2018}
H.~Shim, S.~Hwang, and E.~Yang, ``Joint active feature acquisition and
  classification with variable-size set encoding,'' in \emph{Neural Information
  Processing Systems (NeurIPS)}, 2018.

\bibitem{Joshi_2010}
A.~Joshi, F.~Porikli, and N.~Papanikolopoulos, ``Breaking the interactive
  bottleneck in multi-class classification with active selection and binary
  feedback,'' in \emph{IEEE Conference on Computer Vision and Pattern
  Recognition (CVPR)}, 2010.

\bibitem{Biswas_2012}
A.~Biswas and D.~Jacobs, ``Active image clustering: Seeking constraints from
  humans to complement algorithms,'' in \emph{IEEE Conference on Computer
  Vision and Pattern Recognition (CVPR)}, 2012.

\bibitem{Xiong_2015}
S.~Xiong, Y.~Pei, R.~Rosales, and X.~Fern, ``Active learning from relative
  comparisons,'' \emph{IEEE Transactions on Knowledge and Data Engineering},
  vol.~27, no.~12, 2015.

\bibitem{Qian_2013}
B.~Qian, X.~Wang, F.~Wang, H.~Li, J.~Ye, and I.~Davidson, ``Active learning
  from relative queries,'' in \emph{International Joint Conference on
  Artificial Intelligence (IJCAI)}, 2013.

\bibitem{Aditya_2019_WACV}
A.~Bhattacharya and S.~Chakraborty, ``Active learning with n-ary queries for
  image recognition,'' in \emph{IEEE Winter Conference on Applications of
  Computer Vision (WACV)}, 2019.

\bibitem{Joshi_2012}
A.~Joshi, F.~Porikli, and N.~Papanikolopoulos, ``Scalable active learning for
  multiclass image classification,'' \emph{IEEE Transactions on Pattern
  Analysis and Machine Intelligence (TPAMI)}, vol.~34, no.~11, pp. 2259 --
  2273, 2012.

\bibitem{ECML_Paper}
T.~Sabata, P.~Pulc, and M.~Holena, ``Semi-supervised and active learning in
  video scene classification from statistical features,'' in \emph{Workshop at
  the European Conference on Machine Learning (ECML)}, 2018.

\bibitem{Transportation_Journal}
S.~Sivaraman and M.~Trivedi, ``A general active-learning framework for on-road
  vehicle recognition and tracking,'' \emph{IEEE Transactions on Intelligent
  Transportation Systems (TITS)}, vol.~11, no.~2, pp. 267 -- 276, 2010.

\bibitem{Yan_2003}
R.~Yan, J.~Yang, and A.~Hauptmann, ``Automatically labeling video data using
  multi-class active learning,'' in \emph{IEEE International Conference on
  Computer Vision (ICCV)}, 2003.

\bibitem{SVM_AL_1}
S.~Vijayanarasimhan, P.~Jain, and K.~Grauman, ``Far-sighted active learning on
  a budget for image and video recognition,'' in \emph{IEEE Conference on
  Computer Vision and Pattern Recognition (CVPR)}, 2010.

\bibitem{SVM_AL_2}
L.~Zhao, G.~Sukthankar, and R.~Sukthankar, ``Robust active learning using
  crowdsourced annotations for activity recognition,'' in \emph{Workshop at the
  AAAI Conference on Artificial Intelligence}, 2011.

\bibitem{Bandla_2013}
S.~Bandla and K.~Grauman, ``Active learning of an action detector from
  untrimmed videos,'' in \emph{IEEE International Conference on Computer Vision
  (ICCV)}, 2013.

\bibitem{Contrastive_AL}
S.~Ma, Z.~Zeng, D.~McDuff, and Y.~Song, ``Active contrastive learning of
  audio-visual video representations,'' in \emph{International Conference on
  Learning Representations (ICLR)}, 2021.

\bibitem{AL_Video_Tracking}
S.~Behpour, ``Active learning in video tracking,'' in \emph{arXiv:1912.12557},
  2020.

\bibitem{AL_Video_Description}
D.~Chan, S.~Vijayanarasimhan, D.~Ross, and J.~Canny, ``Active learning for
  video description with cluster-regularized ensemble ranking,'' in \emph{Asian
  Conference on Computer Vision (ACCV)}, 2020.

\bibitem{AL_Video_Recommendation}
J.~Cai, J.~Tang, Q.~Chen, Y.~Hu, X.~Wang, and S.~Huang, ``Multi-view active
  learning for video recommendation,'' in \emph{International Joint Conference
  on Artificial Intelligence (IJCAI)}, 2019.

\bibitem{AL_Video_Segmentation}
A.~Fathi, M.~Balcan, X.~Ren, and J.~Rehg, ``Combining self training and active
  learning for video segmentation,'' in \emph{British Machine Vision Conference
  (BMVC)}, 2011.

\bibitem{Shen_2004}
D.~Shen, J.~Zhang, J.~Su, G.~Zhou, and C.~Tan, ``Multi-criteria based active
  learning for named entity recognition,'' in \emph{Association for
  Computational Linguistics (ACL)}, 2004.

\bibitem{Sriperumbudur_2011}
B.~Sriperumbudur, K.~Fukumizu, and G.~Lanckriet, ``Universality, characteristic
  kernels and rkhs embedding of measures,'' \emph{Journal of Machine Learning
  Research (JMLR)}, vol.~12, 2011.

\bibitem{Yuan_JMLR_2013}
X.~Yuan and T.~Zhang, ``Truncated power method for sparse eigenvalue
  problems,'' \emph{Journal of Machine Learning Research (JMLR)}, vol.~14, pp.
  899 -- 925, 2013.

\bibitem{Random_Proj}
W.~Johnson and J.~Lindenstrauss, ``Extensions of lipschitz mappings into a
  hilbert space,'' in \emph{Conference in Modern Analysis and Probability},
  1984.

\bibitem{Vempala_2004}
S.~Vempala, ``The random projection method,'' in \emph{Americal Mathematical
  Society}, 2004.

\bibitem{Reza_2009}
R.~Farahani and M.~Hekmatfar, \emph{Facility Location: Concepts, Models,
  Algorithms and Case Studies}.\hskip 1em plus 0.5em minus 0.4em\relax
  Physica-Verlag HD, 2009.

\bibitem{Cook_1998}
W.~Cook, W.~Cunningham, W.~Pulleyblank, and A.~Schrijver, \emph{Combinatorial
  Optimization}.\hskip 1em plus 0.5em minus 0.4em\relax Springer, 1998.

\bibitem{Soomro_TR2012}
K.~Soomro, A.~Zamir, and M.~Shah, ``Ucf 101: A dataset of 101 human action
  classes from videos in the wild,'' in \emph{Techical Report, UCF}, 2012.

\bibitem{Kinetics_dataset}
W.~Kay, J.~Carreira, K.~Simonyan, B.~Zhang, C.~Hillier, S.~Vijayanarasimhan,
  F.~Viola, T.~Green, T.~Back, P.~Natsev, M.~Suleyman, and A.~Zisserman, ``The
  kinetics human action video dataset,'' in \emph{arXiv:1705.06950}, 2017.

\end{thebibliography}

\end{document}